\documentclass[letterpaper, 10 pt, conference]{ieeeconf}  % Comment this line out if you need a4paper

\IEEEoverridecommandlockouts                              % This command is only needed if
\usepackage{amsthm}
\usepackage{amssymb}  % assumes amsmath package installed

\usepackage{graphics} % for pdf, bitmapped graphics files
\usepackage{float,url,enumerate}
\usepackage{graphicx}
\usepackage{hyperref}

\usepackage{algorithm}
\usepackage{algorithmic}

\usepackage[usenames, dvipsnames]{color} % for comments

\usepackage{xspace}
\makeatletter
\DeclareRobustCommand\onedot{\futurelet\@let@token\@onedot}
\def\@onedot{\ifx\@let@token.\else.\null\fi\xspace}
\def\eg{\emph{e.g}\onedot} 
\def\ie{\emph{i.e}\onedot} 
 
\def\etc{\emph{etc}\onedot} 
 
\def\etal{\emph{et al}\onedot}
\makeatother

\newcommand{\C}{\mathcal{C}} % special contact frame

\newcommand{\G}{{\bf G}} % set of grasp points
\title{\vspace{-0cm}
\LARGE \bf
Robust Execution of Contact-Rich Motion Plans by Hybrid Force-Velocity Control
}

\author{Yifan Hou and Matthew T. Mason~\IEEEmembership{Fellow,~IEEE} % <-this % stops a space
\thanks{*This work was supported under NSF Grant No. 1662682. }% <-this % stops a space
\thanks{The authors are with the Robotics Institute, Carnegie Mellon University, Pittsburgh, PA 15213, USA.
        {\tt\small yifanh@cmu.edu, matt.mason@cs.cmu.edu}}
}

\begin{document}

\maketitle
\thispagestyle{empty}
\pagestyle{empty}

%%%%%%%%%%%%%%%%%%%%%%%%%%%%%%%%%%%%%%%%%%%%%%%%%%%%%%%%%%%%%%%%%%%%%%%%%%%%%%%%
\begin{abstract}
In hybrid force-velocity control, the robot can use velocity control in some directions to follow a trajectory, while performing force control in other directions to maintain contacts with the environment regardless of positional errors. We call this way of executing a trajectory \textit{hybrid servoing}.
We propose an algorithm to compute hybrid force-velocity control actions for hybrid servoing. We quantify the robustness of a control action and make trade-offs between different requirements by formulating the control synthesis as optimization problems.
Our method can efficiently compute the dimensions, directions and magnitudes of force and velocity controls.
We demonstrated by experiments the effectiveness of our method in several contact-rich manipulation tasks.
Link to the video: \href{https://youtu.be/KtSNmvwOenM}{https://youtu.be/KtSNmvwOenM}.

\end{abstract}

%%%%%%%%%%%%%%%%%%%%%%%%%%%%%%%%%%%%%%%%%%%%%%%%%%%%%%%%%%%%%%%%%%%%%%%%%%%%%%%%
%%%%%%%%%%%%%%%%%%%%%%%%%%%%%%%%%%%%%%%%%%%%%%%%%%%%%%%%%%%%%%%%%%%%%%%%%%%%%%%%

% !TEX root = ../ICRA19Hybrid.tex

\section{INTRODUCTION}
\label{sec:intro}

% example
In the materials handling industry where robots pick up random objects from bins, it's generally difficult to pick up the last few objects, because they are usually too close to the bin walls, leaving no collision-free grasp locations. It's even harder if a flat object is lying in the corner. However, in such cases a human would simply lift the object up with only one finger by pressing on a side of the object and pushing against the bin wall.
% manipulation with contacts is useful
This is one of the many examples where humans can solve manipulation problems that are difficult for robots with surprisingly concise solutions. The human finger can do more than the robot finger because the human naturally utilizes the contacts between the object and the environment to create solutions.

Manipulation under external contacts is common and useful in human life, yet our robots are still far less capable of doing it than they should be.
In the robot motion planning community, most works are focused on generating collision-free motion trajectories.
There are planning methods that are capable of computing complicated, contact-rich robot motions \cite{mordatch2012discovery,posa2014direct}, however, the translation from a planned motion to a successful experiment turns out to be difficult. High stiffness servo controls, such as velocity control, are prone to positional errors in the model. Low stiffness controls such as force control are vulnerable to all kinds of inevitable force disturbances and noise, such as un-modeled friction.

In this work, we attempt to close the gap between contact-rich motion planning and successful execution with \textit{hybrid servoing}, defined as using hybrid force-velocity control to execute the planned trajectory. We try to combine the good points of both worlds: high stiffness controls are immune to small force disturbances, while force controls (even somewhat inaccurate force controls) can comply with holonomic constraints under modeling uncertainties.

Solving for hybrid force-velocity control is more difficult than solving for force or velocity alone, because we need to compute directions for each type of control. It is challenging to properly formulate the problem itself; the solution space is also higher dimensional. This is why most of the previous works on hybrid force-velocity control only analyzed simple systems with the robot itself (may include a firmly grasped object) and a rigid environment, without any free objects and more degree-of-freedoms.

In this work, we provide a hybrid servoing problem formulation that works for systems with more objects, along with an algorithm to efficiently solve it.
We quantify what it means for a constraint to be satisfied ``robustly'', and automate the control synthesis by formulating it as two optimization problems on the velocity/force controlled actions. The optimization automatically makes trade-offs between robustness and feasibility.
In particular, we show that the velocity controlled directions do not have to be orthogonal to the holonomic constraints, leaving space for more solutions. Being closer to orthogonal does have benefits; it is considered in the cost function.

The rest of the paper is organized as follows. In the next section we review the related works. In section \ref{sec:problem_formulation}, we introduce our modeling and problem formulation for hybrid servoing. In section \ref{sec:approach}, we describe our algorithm for solving hybrid servoing. In section \ref{sec:example} and \ref{sec:experiments}, we provide a step by step analysis for one simple example, along with experimental results for several examples.

% Our algorithm can be solved efficiently. The most computational

% !TEX root = ../ICRA19Hybrid.tex

\section{RELATED WORK}
\label{sec:related_work}

\subsection{Hybrid Force-Velocity Control} % (fold)
\label{sub:hybrid_force_velocity_control}
The idea of using hybrid force-velocity control for manipulation under constraints can date back to 1980s.
Mason \cite{mason1981compliance} introduced a framework for identifying force and velocity controlled directions in a task frame given a task description. Raibert and Craig \cite{raibert1981hybrid} completed the framework and demonstrated a working system.
Yoshikawa \cite{yoshikawa1987dynamic} investigated hybrid force-velocity control in joint space under Cartesian space constraints, and proposed to use gradient of the constraints to find the normal of the constraint surface in the robot joint space.
There are also works on modeling the whole constrained robot system using Lagrange dynamics, such as analyzing the system stability under hybrid force-velocity control \cite{mcclamroch1988feedback}, or performing Cartesian space tracking for both positions and forces \cite{mills1989force}. Most of these works modeled only the robot and a rigid environment without any un-actuated degree-of-freedoms in the system. As an exception, Uchiyama and Dauchez performed hybrid force-velocity control for a particular example: two manipulators contacting one object \cite{uchiyama1988symmetric}.

There are lots of works on how to implement hybrid force-velocity controls on manipulators. For example, stiffness control can be used for this purpose. Velocity control is essentially a high stiffness control; force control can be implemented by low stiffness control with force offset.
Salisbury \cite{salisbury1980active} described how to perform stiffness control on arbitrary Cartesian axes with a torque-controlled robot.
Raibert and Craig \cite{raibert1981hybrid} divided Cartesian space into force/velocity controlled parts, then controlled them with separated controllers.
The impedance control \cite{hogan1985impedance}  and operational space control \cite{khatib1987unified} theory provided detailed analysis for regulating the force related behaviors of the end-effector for torque-controlled robots. Maples and Becker described how to use a robot with position controlled inner loop and a wrist-mounted force-torque sensor to do stiffness control on Cartesian axes \cite{maples1986experiments}. Lopes and Almeida enhanced the impedance control performance of industrial manipulators by mounting a high frequency 6DOF wrist \cite{lopes2008force}. Whitney \cite{whitney1987historical} and De Schutter \cite{de1998force} provided overviews and comparisons for a variety of force control methods.

\subsection{Motion Planning through Contacts} % (fold)
\label{sub:motion_planning_through_contacts}
Recently, a lot of works tried to solve manipulation under constraints without explicitly using force control. For holonomic constraints, De Schutter \etal proposed a constraint-based motion planning and state estimation framework \cite{de2007constraint}. Berenson \etal did motion planning on the reduced manifold of the constrained state space \cite{berenson2009manipulation}. For non-holonomic constraints, the most popular example is pushing \cite{lynch-mason-pushing,zhou2016convex,dogar2011framework}. Chavan-Dafle \etal performed in-hand manipulation by pushing the object against external contacts \cite{prehensile}. In these works, the robots interacted with the objects in a way that force control was not necessary.

% !TEX root = ../ICRA19Hybrid.tex

\section{MODELING \& PROBLEM FORMULATION}
\label{sec:problem_formulation}
First of all, we introduce how we model a hybrid servoing problem. We adopt quasi-static assumption throughout the work, \ie inertia force and Coriolis force are negligible. All objects and the robot are rigid. A motion trajectory is available such that the goal for our algorithm at any time step can be given as instantaneous velocities. All analysis in this section and the next section are conducted for one time step.
For consistency with previous works, we reuse several concepts from \cite{mason1981compliance} such as \textit{natural constraints} and \textit{artificial constraints}. To better suit a more general problem formulation, we extend the meanings of these terms when necessary, much to the second author's consternation.

\subsection{Symbols} % (fold)
\label{sub:symbols}
Consider a system of rigid bodies including the robot and at least one object. Denote $q\in \mathbb{R}^{n_q}$ as the configuration of the whole system. Denote $\tau\in \mathbb{R}^{n_q}$ as the corresponding force variable (internal forces), \ie if $q$ denotes joint angles, $\tau$ denotes joint torques. Although the configuration space is enough to encode the state of the system, its time derivative may not make sense as a velocity, \eg when $q$ contains quaternions. We describe the system velocity in a different space, the selection of the variables is usually called the ``generalized variables''.

Denote $v=[v^T_u\  v^T_a]^T \in \mathbb{R}^{n}$ as the generalized velocity. We pick the variables of $v$ in such an order that the first $n_u$ elements $v_u\in\mathbb{R}^{n_u}$ denote the uncontrolled (free) dimensions in the system, such as the velocity of an object; the last $n_a$ elements $v_a\in\mathbb{R}^{n_a}$ represent the degrees-of-freedom of the robot actuation.
$v\ne \dot q$ in general, but is related to $\dot q$ by a linear transformation:  $\dot q=\Omega(q)v$, where $\Omega(q)\in \mathbb{R}^{n_q\times n}$.
Denote $f=[f^T_u\ f^T_a]^T\in\mathbb{R}^{n}$ as the generalized force vector (the internal force). The product of $f$ and $v$ is the work done by the robot. Note that the uncontrolled part of $f$ is always zero: $f_u=0$. In the following we will do most of our analysis in the language of generalized variables.

\subsection{Goal Description} % (fold)
\label{sub:goal_description}
The goal for our control at a time step is an affine constraint on the generalized velocity:
\begin{equation}
\label{eq:goal}
	Gv=b_G.
\end{equation}
The goal (\ref{eq:goal}) could be a desired generalized velocity, in which case $G=I$.
The goal may also only involve some entries of $v$. For example, in regrasping problems people only care about the in-hand pose of the object; the pose of the hand can be set free to allow for more solutions.
If a motion trajectory is available, the desired velocity can be obtained from its time derivatives.

% subsection symbols (end)
\subsection{Natural Constraints} % (fold)
\label{sub:natural_constraints}
The law of physics constrains the system in many ways. These constraints will never be violated, no matter what actions the robot takes. We call them the \textit{natural constraints}.  Our definition of the natural constraints includes holonomic constraints and the Newton's second law. The original definition in \cite{mason1981compliance} did not contain the Newton's second law, because it is of no significance for fully actuated systems.
\subsubsection{Holonomic Constraints} % (fold)
\label{subsub:holonomic_constraints}
Holonomic constraints are bilateral constraints on $q$ that are also independent of $\dot q$. Examples are persistent contact constraints and sticking contact constraints. We describe them by
\begin{equation}
\label{eq:holonomic_constraints}
	\Phi(q)=0,
\end{equation}
where $\Phi(q)\in \mathbb{R}^{n_\Phi}$. Its time-derivative gives the constraint on instantaneous velocity:
\begin{equation}
\label{eq: holonomic_constraints_on_velocity}
	J_\Phi(q)\dot q=J_\Phi(q)\Omega(q)v=0.
\end{equation}
If an action attempts to violate a holonomic constraint, \eg pressing an object against a table, a reaction force will emerge to maintain the constraint.
Denote $\lambda\in \mathbb{R}^{n_\Phi}$ as the reaction forces for $\Phi(q)$.
Its positive direction is determined by how we define $\Phi(q)$: when both $\delta\Phi$ and reaction force $\lambda$ are positive, they make positive work. Be careful when applying the rule to the contact forces between two movable objects, as the force would have a different direction for each body.
The contribution of $\lambda$ to the joint torque can be computed from the principle of virtual work \cite{villani2008force}:
$$\tau_\lambda=J^T_\Phi(q)\lambda,$$
project $\tau_\lambda$ into the space of generalized force:
\begin{equation}
	f_\lambda=\Omega^T(q)\tau_\lambda=\Omega^T(q)J^T_\Phi(q)\lambda
\end{equation}
\subsubsection{Newton's second law} % (fold)
\label{subsub:newton_s_second_law}
For systems that are not fully-actuated, Newton's second law becomes necessary.
Denote $F \in \mathbb{R}^{n}$ as the external force (gravity, magnetic force, \etc) in the generalized force coordinates.
Newton's second law requires the sum of all the forces in the system to be zero:
\begin{equation}
\label{eq:newton_second_law}
	 \Omega^T(q)J^T_\Phi(q)\lambda + f + F = 0
\end{equation}
The three terms are contact reaction forces, control actions (internal forces) and external forces, respectively.

\subsection{Velocity controlled actions and holonomic constraints} % (fold)
\label{sub:velocity_controlled_actions_and_holonomic_constraints}
\begin{figure}[h]
    \centering
    \includegraphics[width=0.4\textwidth]{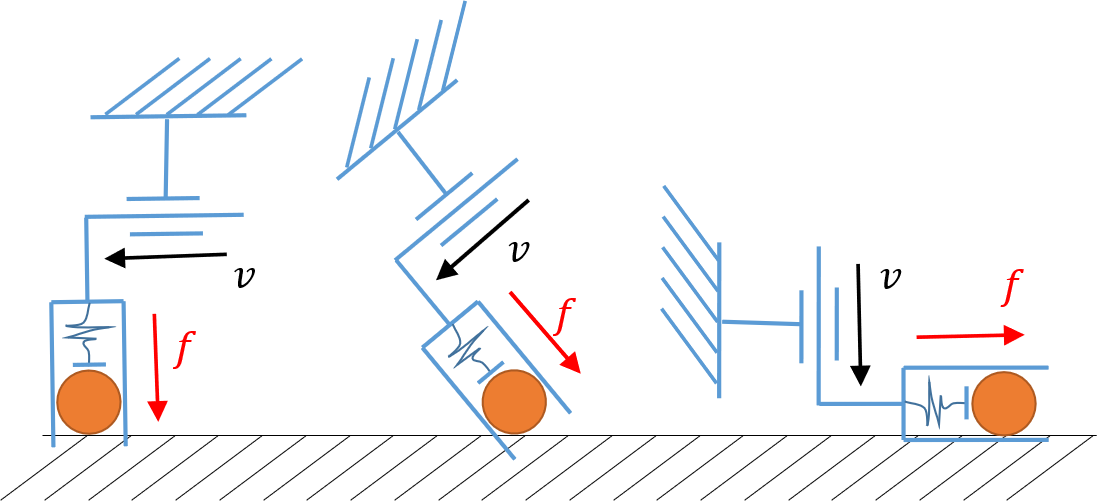}
    \caption{Relation between velocity commands and holonomic natural constraints. The robot (blue) has a velocity controlled joint and a force controlled joint, which are orthogonal to each other. The table provides a natural constraint that stops the object from moving down. Assume no collision between the robot and the table. Systems in the left and middle are feasible. The right system is infeasible.}
    \label{fig:example_1}
\end{figure}
In some works of quasi-static analysis, the rows of Newton's second law for velocity controlled dimensions are ignored because the forces have no influence on other parts of the system, as shown in Fig. \ref{fig:example_1}, left. We keep these rows in (\ref{eq:newton_second_law}), because the axes of velocity commands may not lie completely in the null space of natural constraints, then the force generated from a velocity command would matter in the force computation of the system. One such system is illustrated in Fig. \ref{fig:example_1}, middle.

An interesting question is, can we set velocity commands in any directions? One apparent fact is that the velocity action must not fight against the natural constraints, e.g. trying to push against a wall (Fig. \ref{fig:example_1}, right). Mathematically it means the system of linear equations formed by the natural constraints and velocity commands is infeasible.
The velocity controlled directions are thus preferred to overlap less with the natural constraints, more with its null space, so that the system of equations will be less likely to become infeasible under disturbances. If the system is holonomic, \ie fully actualized, the velocity commands can be chosen from within the null space of the natural constraints \cite{yoshikawa1987dynamic}. This is not always possible in general.
\begin{figure}[h]
    \centering
    \includegraphics[width=0.4\textwidth]{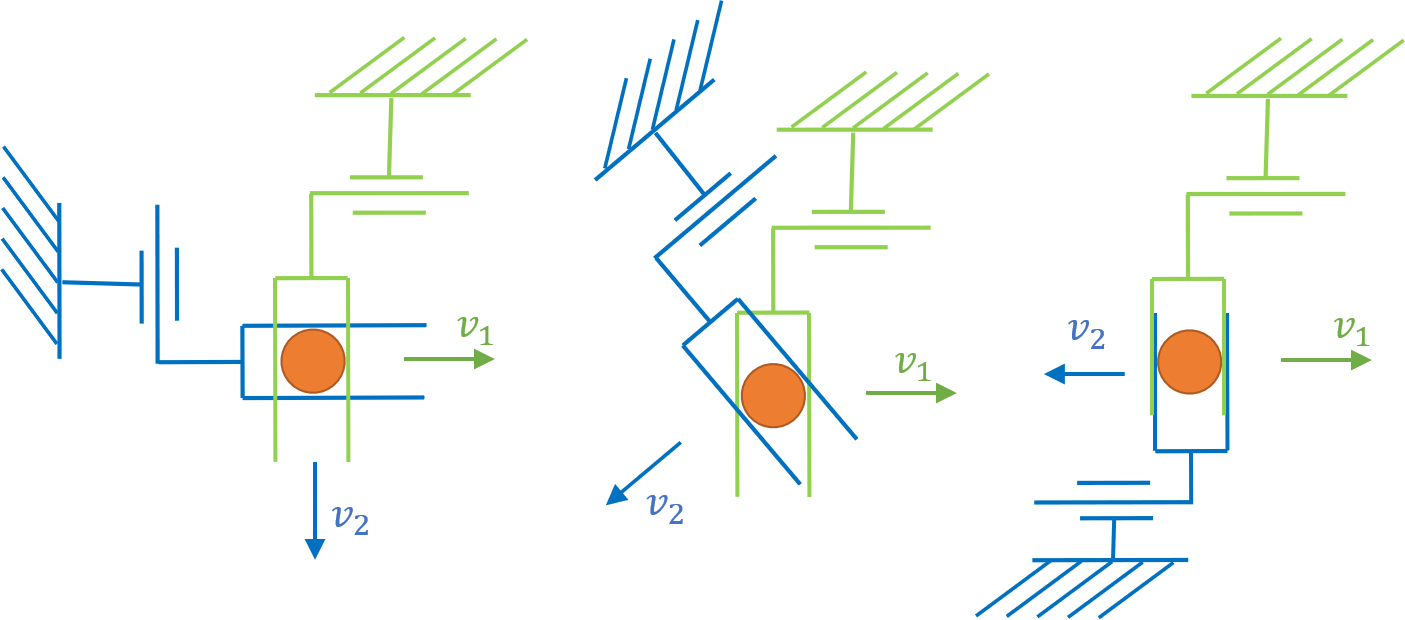}
    \caption{Relation between different velocity commands. The blue robot and green robot are applying different velocity commands on the object. Assume no collision between the two robots. Systems in the left and middle figures are feasible. The right system is infeasible.}
    \label{fig:example_2}
\end{figure}

We can do the same analysis among different velocity commands. As shown in Fig. \ref{fig:example_2}, different velocity controlled directions in generalized velocity space should not be co-linear. The system would be more robust to disturbances if the velocities are more perpendicular to each other.
% subsection goal_description (end)
\subsection{Guard Conditions} % (fold)
\label{sub:guard_conditions}
A contact may be in one of three modes: sliding, sticking or not contacting. A motion plan usually assumes a certain contact mode for each contact at any given time.
In hybrid control theory, the term \textit{guard conditions} refers to conditions for transitions between discrete modes. In our problem, we also need to apply guard conditions to make sure our robot action will maintain the contact modes in the motion plan.

In this work, we consider guard conditions that can be expressed as linear (or affine) constraints on force variables. Examples of this type are friction cone constraints and lower/upper bounds on forces.
\begin{equation}
\label{eq:guard_conditions}
	\Lambda\left[ {\begin{array}{*{20}{c}}
	\lambda\\
	f
	\end{array}} \right] \le {b_\Lambda},\ \ \ \ \Gamma\left[ {\begin{array}{*{20}{c}}
    \lambda\\
    f
    \end{array}} \right] = {b_\Gamma}.
\end{equation}

\subsection{Problem Formulation} % (fold)
\label{sub:problem_formulation}
To clearly describe the actions, we introduce \textit{transformed generalized velocity} $w=[w^T_u\ w^T_{af}\ w^T_{av}]^T\in\mathbb{R}^{n}$, where $w_u=v_u$ is the un-actuated velocity, $w_{af}\in \mathbb{R}^{n_{af}}$ is the velocity in the force controlled directions, $w_{av}\in\mathbb{R}^{n_{av}}$ is the velocity controlled actions. Denote $\eta=[\eta^T_u\ \eta^T_{af}\ \eta^T_{av}]^T\in\mathbb{R}^{n}$ as the \textit{transformed generalized force}, where $\eta_u=f_u=0$ is the un-actuated force, $\eta_{af}\in \mathbb{R}^{n_{af}}$ is the force controlled actions, $\eta_{av}\in\mathbb{R}^{n_{av}}$ is the force in the velocity controlled directions. The action space of the robot is $(w_{av},\eta_{af})$.

We use matrix $T$ to describe the directions of force/velocity controlled axes: $w=Tv$, $\eta=Tf$.
$T=diag(I_u, R_a)\in \mathbb{R}^{n\times n}$, where $I_u\in \mathbb{R}^{n_u\times n_u}$ is an identity matrix, $R_a\in \mathbb{R}^{n_a\times n_a}$ is an invertible matrix (not necessarily orthogonal).
Now we are ready to define the hybrid servoing problem mathematically. At any time step during the execution of a motion plan, the task of hybrid servoing is to find out:
\begin{enumerate}
	\item the dimensions of force controlled actions and velocity controlled actions, $n_{af}$ and $n_{av}$, and
	\item the directions to do force control and velocity control, described by the matrix $T$, and
	\item the magnitude of force/velocity actions: $\eta_{af}$ and $w_{av}$,
\end{enumerate}
such that:
\begin{itemize}
	\item the goal (\ref{eq:goal}) is satisfied as a result of velocity controlled actions and holonomic constraints (\ref{eq:holonomic_constraints});
	\item the guard conditions (\ref{eq:guard_conditions}) are satisfied as a result of force controlled actions and the Newton's law (\ref{eq:newton_second_law}).
\end{itemize}
Usually the problem described above has more than one solution. As discussed in section \ref{sub:velocity_controlled_actions_and_holonomic_constraints}, we prefer velocity commands that are perpendicular to each other, and are close to the null space of holonomic natural constraints.

Under this formulation, the satisfaction of goals is ensured by velocity controlled actions, which are accurate and immune to force disturbances; the holonomic natural constraints are satisfied by selecting non-conflicting directions for velocity controlled actions, it won't be easy for a disturbance to make them conflict again. The guard conditions are basically maintaining contacts, which do not require the force controlled actions to be super precise. These are the keys to the robustness of our method.

% subsection problem_formulation (end)

% !TEX root = ../ICRA19Hybrid.tex

\section{APPROACH}
\label{sec:approach}
% ===============================================================
Now we introduce an algorithm to efficiently solve the problem defined in section \ref{sub:problem_formulation}. The algorithm first solves for velocity commands, during which the dimensions and directions of both velocity control and force control are also determined. Then we fix the directions and solve for force controlled actions.

\subsection{Solve for Velocity Controlled Actions} % (fold)
\label{sub:solve_for_velocity_controlled_actions}
In this section, we design the velocity command (solve for $n_{af}, n_{av}, T$ and $w_{av}$), so as to satisfy all the velocity-level conditions.
We use a $n_{av}\times n$ selection matrix $S_{av}$ to select the velocity commands out of the generalized variables: $w_{av}=S_{av}w$. Equations of interest to this section are:
(Use $\dot q=\Omega v, w=Tv$, omitting argument $q$)
\begin{itemize}
  \item Holonomic natural constraint $J_\Phi\Omega v=0$. Denote $N=J_\Phi\Omega$, the constraint becomes $Nv=0$;
  \item Goal condition $Gv=b_G$;
  \item Velocity command $S_{av}Tv=w_{av}$. Denote $C=S_{av}T$, $b_C=w_{av}$, rewrite the velocity command as $Cv=b_C$.
\end{itemize}
Denote the solution set of each equation as $Sol(N), Sol(G)$ and $Sol(C)$. We need to design the velocity command $C,\ b_C$ such that the resulted solution space (the solution set of natural constraints and velocity commands) becomes \textit{a non-empty subset of the desired generalized velocities} (the solution set of natural constraints and goal condition):
\begin{equation}
Sol(N\&C)\ \in\ Sol(N\&G)
\end{equation}

\subsubsection{Determine dimensions of velocity control} % (fold)
\label{subsub:determine_dimensions_of_velocity_control}
Denote $r_N={\rm rank}(N), r_{NG}={\rm rank}(\left[ {\begin{array}{*{20}{c}}{N }\\G\end{array}}\right])$. The minimum number of independent velocity control we must enforce is
\begin{equation}
\label{eq:nav_min}
  {n^{\min}_{av}} = r_{NG}-r_N.
\end{equation}
This condition makes sure the dimension of $Sol(N\&C)$ is smaller or equal to the dimension of $Sol(N\&G)$, so that their containing relationship becomes possible. Physically it means the velocity commands need to reduce enough degree-of-freedoms from the null space of the natural constraints. The maximum number of independent velocity commands we can enforce is
\begin{equation}
\label{eq:nav_max}
  n^{\max}_{av} = n - r_N={\rm Dim}({\rm null}(N)),
\end{equation}
where ${\rm null}(N)$ denotes the null space of $N$. This condition ensures the system will not be overly constrained to have no solution.
We choose the minimal number of necessary velocity constraints:
\begin{equation}
\label{eq:choice_of_nav}
n_{av}=n^{\min}_{av} =r_{NG}-r_N.
\end{equation}
This choice makes it easier for the system to avoid jamming. As will be shown in the next section, it also leaves more space for solving force controlled actions.

\subsubsection{Solve for directions and magnitude} % (fold)
\label{subsub:solve_for_directions_and_magnitude}
With our choice of $n_{av}$, we know ${\rm rank}([N; C])={\rm rank}([N; G])$. Then the condition $Sol(N\&C)\ \in\ Sol(N\&G)$ implies
\begin{equation}
\label{eq:sol_NC_equals_sol_NG}
  Sol(N\&C)\ =\ Sol(N\&G),
\end{equation}
\ie the two linear systems share the same solution space. (\ref{eq:sol_NC_equals_sol_NG}) can be achieved by firstly choosing $C$ such that the homogeneous linear systems $\left[ {\begin{array}{*{20}{c}}N\\C\end{array}} \right]v = 0$ and $\left[ {\begin{array}{*{20}{c}}N\\G\end{array}} \right]v = 0$
become equivalent (share the same solution space). Compute a basis for the solution of $[N^T G^T]^Tv=0$: $[\sigma_1,...,\sigma_{n-r_{NG} }]$, then we just need to ensure $C$ satisfies:
\begin{equation}
\label{eq:constraints_on_C}
  C\sigma_i=0,\ \  i=1,...,n-r_{NG}
\end{equation}
Then we can compute $b_C$ from any specific solution of $\{Nv=0, Gv=b_G\}$. The original non-homogeneous systems then become equivalent.

Beside equation (\ref{eq:constraints_on_C}), we have a few more requirements/preferences on $C$ based on the discussions in section \ref{sub:velocity_controlled_actions_and_holonomic_constraints}:
\begin{itemize}
  \item Rows of $C$ must be linearly independent from each other. And we prefer to have them as orthogonal to each other as possible.
  \item Each row of $C$ is also independent from rows of the holonomic natural constraint $N$. We prefer to pick the rows as close to ${\rm null}(N)$ as possible.
\end{itemize}
To solve for $C$, denote $c^T\in \mathbb{R}^{1\times n}$ as any row in $C$. From $C=S_{av}T$ we know the first $n_u$ columns in $C$ are zeros, rewrite this and equation (\ref{eq:constraints_on_C}) as a linear constraint on $c$:
\begin{equation}
\label{eq:constraint_on_c}
  \left[ {\begin{array}{*{20}{c}}{\begin{array}{*{20}{c}}{\sigma _1^T}\\ \vdots \\{\sigma _{{n} - {n_{NG}}}^T}\end{array}}\\{\left[ {\begin{array}{*{20}{c}}{{I_{{n_u}}}}&{{{\bf{0}}_{{n_u} \times {n_a}}}}
  \end{array}} \right]}\end{array}} \right]c  = \left[ {\begin{array}{*{20}{c}}
  0\\ \vdots \\0\end{array}} \right]
\end{equation}
Its solution space has dimension of $n_c = n_a - n + r_{NG} = r_{NG} - n_u$. Since we need $n_{av}$ independent constraints, we require $n_c=r_{NG}-n_u \ge n_{av}=r_{NG}-r_N$, which gives $r_N\ge n_u $, \ie
\begin{equation}
\label{eq:dimensional_requirement}
r_N+n_a\ge n.
\end{equation}
For our method to work, (\ref{eq:dimensional_requirement}) says it must be possible for the actions and constraints to fully constrain the system.
Denote matrix $\mathbb{B}_c=[c^{(1)}\ \cdots\ c^{(n_c)}]$ as a basis of the solution space of equation (\ref{eq:constraint_on_c}). Denote ${\rm Null}(N)$ as a basis of ${\rm null}(N)$.
We can find a $C$ that satisfies all the conditions by solving the following optimization problem:
\begin{equation}
\label{eq:optimization_for_velocity}
\begin{array}{l}
\mathop {\min }\limits_{{{\bf{k}}_1}, \cdots ,{{\bf{k}}_{{n_{av}}}}} \sum\limits_{i \ne j} {||{c_i^T}c _j|{|}}  - \sum\limits_i {||{\rm Null}(N)^T c _i|{|}} \\
s.t.\;\;\,\;\;\,{c_i^T}c_i = 1,\;\;\,\forall i\\
\;\;\,\;\;\,\;\;\,\;\;\,{c _i} = \mathbb{B}_c{{\bf{k}}_i},\;\;\,\forall i
\end{array}
\end{equation}
The best velocity constraint $C^*=(\mathbb{B}_c\;[{\bf k_1\;\;\;...\;\; k_{n_{av}}}])^T$. The optimization problem (\ref{eq:optimization_for_velocity}) is non-convex because of the unit length constraint $c_i^Tc_i=1$. However, we can solve the problem numerically by projecting the solution back to the constraint after each gradient update:
\begin{enumerate}
  \item Start from a random ${\bf k=[{\bf k_1\;\;\;...\;\; k_{n_{av}}}]}$;
  \item Perform a gradient descent step: ${\bf k}\gets {\bf k} - t\nabla f$;
  \item Projection: ${\bf k_i}\gets \frac{\bf k_i}{||\mathbb{B}_c{\bf k_i}||},\;\;\;\;\forall i$;
  \item Repeat from step two until convergence.
\end{enumerate}
Here $\nabla f$ is the gradient of the cost function reshaped to the same size as $\bf k$. $t=10$ is a step length. In practice, we run the projected gradient descent algorithm above with $N_s$ different initializations to avoid bad local minima.

After obtaining $C^*$, we know the last $n_{av}$ rows of $R_a$. Denote the last $n_a$ columns of $C^*$ as $R_{C^*}$, we can expand it into a full rank $R_a$:
\begin{equation}
\label{eq:expand_R_a}
{R_a} = \left[ {\begin{array}{*{20}{c}}
{{\rm Null}{{({R_{C^*}})}^T}}\\
{{R_{C^*}}}
\end{array}} \right],
\end{equation}
it encodes the axes of the force controlled directions. Then we have $T=\rm{diag}(I_u, R_a)$.
The procedures are summarized in algorithm \ref{alg:solve_for_velocity}.
\begin{algorithm}[h]
\caption{Solve for velocity controlled actions}  \label{alg:solve_for_velocity}
\begin{algorithmic}[1]
% \renewcommand{\algorithmicrequire} {\textbf{Input :} }
% \REQUIRE $\qobj\ui,\qobj\uf,\qgrp\uf,\P$, number of samples $N_s$.
\STATE Check condition (\ref{eq:dimensional_requirement}) for feasibility.
\STATE Compute $n_{av}$ from equation (\ref{eq:choice_of_nav}).
\STATE Compute a basis of $[N^T G^T]^Tv=0$, plug in equation (\ref{eq:constraint_on_c}) and compute a basis $\mathbb{B}_c$.
\STATE Sample $N_s$ sets of coefficients ${\bf k}\in\mathbb{R}^{n_c\times n_{av}}$
\FOR {each sample $\bf k$}
\STATE Solve the optimization problem (\ref{eq:optimization_for_velocity}).
\STATE Compute $C=(\mathbb{B}_c{\bf k})^T$ from the solution.
\STATE Compute the cost of $C$ from equation (\ref{eq:optimization_for_velocity}).
\ENDFOR
\STATE Pick the $C^*$ with lowest cost.
\STATE Use equation (\ref{eq:expand_R_a}) to compute $R_a$. Then $T=\rm{diag}(I_u, R_a)$.
\STATE Compute one solution $v^*$ for $Nv=0, Gv=b_G$.
\STATE Compute $w_{av}=b_C=C^*v^*$.
\end{algorithmic}
\end{algorithm}

% subsection solve_for_velocity_controlled_actions (end)
\subsection{Solve for Force Controlled Actions} % (fold)
\label{sub:solve_for_force_controlled_actions}
Next we compute the force command (solve for $\eta_{af}$) so as to satisfy all the force-level requirements.
Equations of interest to this section:
(Use $\eta=Tf$, omitting argument $q$)
\begin{itemize}
  \item Newton's second law: express (\ref{eq:newton_second_law}) in the transformed generalized force space:
    \begin{equation}
      \label{eq:newton_second_law_lambda_eta}
      T\Omega^T(q)J^T_\Phi(q)\lambda + \eta + TF = 0.
    \end{equation}
  \item Guard conditions: express them as constraints on $\lambda, \eta$:
    \begin{equation}
      \label{eq:guard_condition_lambda_eta_inequality}
      \Lambda\left[ {\begin{array}{*{20}{c}}\lambda \\f\end{array}} \right] = [{\Lambda_\lambda }\;\;\,{\Lambda_f}{T^{ - 1}}]\left[ {\begin{array}{*{20}{c}}\lambda \\\eta \end{array}} \right] \le {b_\Lambda}.
    \end{equation}
    \begin{equation}
      \label{eq:guard_condition_lambda_eta_equality}
      \Gamma\left[ {\begin{array}{*{20}{c}}\lambda \\f\end{array}} \right] = [{\Gamma_\lambda }\;\;\,{\Gamma_f}{T^{ - 1}}]\left[ {\begin{array}{*{20}{c}}\lambda \\\eta \end{array}} \right] = {b_\Gamma}.
    \end{equation}
\end{itemize}
The unknowns are the force variables $\lambda, \eta$. Remember $\eta=[\eta_u, \eta_{af}, \eta_{av}]$. All the equality constraints ((\ref{eq:newton_second_law_lambda_eta}), (\ref{eq:guard_condition_lambda_eta_equality}) and our choice of $\eta_{af}$) will determine the value of all the forces; we need to make sure the resulted forces satisfy the inequality constraints (\ref{eq:guard_condition_lambda_eta_inequality}).
Remember also $f_u=0$. Express it as $Hf=HT^{-1}\eta=0$. Combine $HT^{-1}\eta=0$, (\ref{eq:newton_second_law_lambda_eta}) and (\ref{eq:guard_condition_lambda_eta_equality}) into one constraint:
\begin{equation}
  \label{eq:newton_all}
  \left[ {\begin{array}{*{20}{c}}{{0}}&{{H}{T^{ - 1}}}\\{T{\Omega ^T}J_\Phi ^T}&I\\\Gamma_\lambda&\Gamma_fT^{-1}\end{array}} \right]\left[ {\begin{array}{*{20}{c}}\lambda \\\eta \end{array}} \right] = \left[ {\begin{array}{*{20}{c}}{\bf{0}}\\{ - TF}\\b_\Gamma\end{array}} \right].
\end{equation}
Due to the limitation of rigid body modeling, the free forces may not have a unique solution given a force action $\eta_{af}$. Denote the free forces as $f_{free} = [\lambda^T\; \eta_u^T\; \eta_{av}^T]^T$, rewrite the constraints (\ref{eq:newton_all}) and move $\eta_{af}$ to the right hand side, we find one solution for $f_{free}$ by penalizing the sum-of-squares norm of the free forces:
\begin{equation}
  \label{eq:QP_newton_law}
  \begin{array}{l}\mathop {\min }\limits_{f_{free}} \;\;\,f_{free}^Tf_{free}\\s.t.\;\;\,\;\;\,{M_{free}}f_{free} = \left[ {\begin{array}{*{20}{c}}
  {\bf{0}}\\{ - TF}\\b_\Gamma\end{array}} \right] - {M_{{\eta _f}}}{\eta _f}.\end{array}
\end{equation}
This is a quadratic programming (QP) problem. Denote $f^*_{free}$ as the dual variables of $f_{free}$, the KKT condition says the solution to the QP can be found by solving the following linear system:
\begin{equation}
  \label{eq:KKT}
  \left[ {\begin{array}{*{20}{c}}{2I}&{M_{free}^T}\\{{M_{free}}}&{\bf 0}\end{array}} \right]\left[ {\begin{array}{*{20}{c}}f_{free}\\f_{free}^*\end{array}} \right] = \left[ {\begin{array}{*{20}{c}}{\bf 0}\\{\left[ {\begin{array}{*{20}{c}}{\bf{0}}\\{ - TF}\end{array}} \right] - {M_{{\eta _f}}}{\eta _f}}\end{array}} \right].
\end{equation}
This linear system uniquely determines the free forces given force action $\eta_{af}$. Rewrite it as
\begin{equation}
  \label{eq:KKT_final}
  \left[ {\begin{array}{*{20}{c}}{2I}&{M_{free}^T}&{\bf{0}}\\{{M_{free}}}&{\bf{0}}&{{M_{{\eta _f}}}}\end{array}} \right]\left[ {\begin{array}{*{20}{c}}f_{free}\\f_{free}^*\\\eta_{af}\end{array}} \right] = \left[ {\begin{array}{*{20}{c}}{\bf{0}}\\{ - TF}\end{array}} \right].
\end{equation}
This linear equation encodes the unique solution for Newton's law. Finally we solve (\ref{eq:KKT_final}) together with guard conditions (\ref{eq:guard_condition_lambda_eta_inequality}) to compute all forces. The procedure is summarized in algorithm \ref{alg:solve_for_force}.
\begin{algorithm}[h]
\caption{Solve for force controlled actions}  \label{alg:solve_for_force}
\begin{algorithmic}[1]
\STATE From Newton's laws, write down $M_{free}, M_{\eta_{af}}$ in (\ref{eq:QP_newton_law}).
\STATE Write down coefficient matrices for equation (\ref{eq:KKT_final}).
\STATE Solve the linear programming problem (\ref{eq:guard_condition_lambda_eta_inequality})(\ref{eq:KKT_final}) for $\eta_{af}$.
\end{algorithmic}
\end{algorithm}

% subsection solve_for_force_controlled_actions (end)
% !TEX root = ../ICRA19Hybrid.tex

\section{Example}
\label{sec:example}
\begin{figure}[h]
    \centering
    \includegraphics[width=0.4\textwidth]{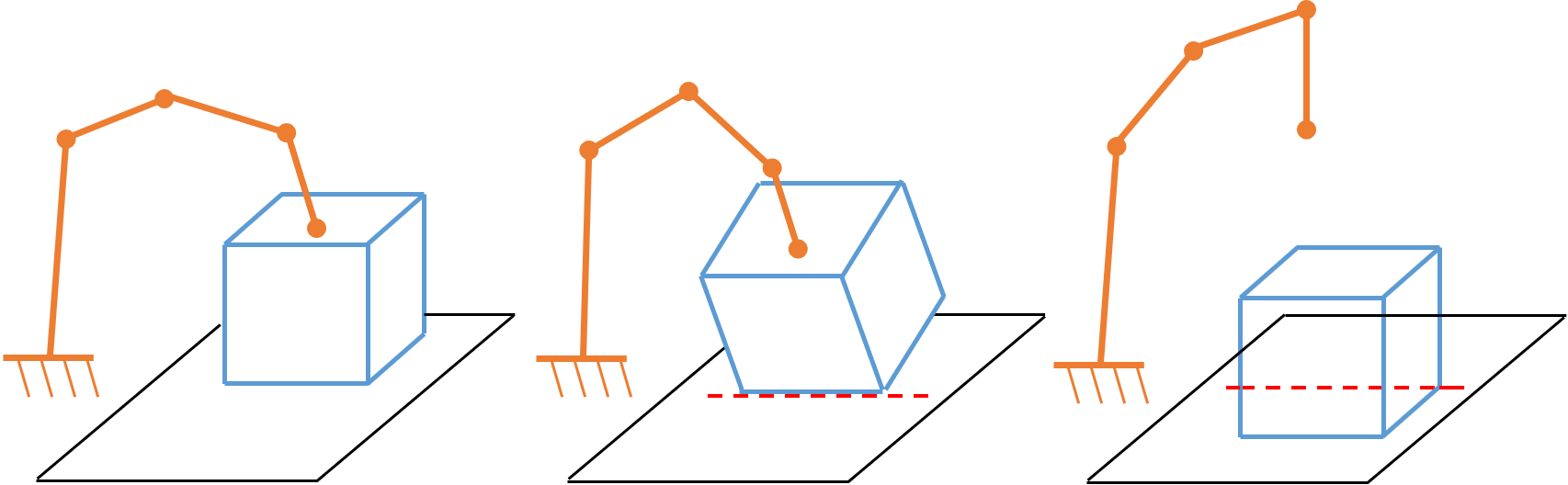}
    \caption{Block tilting example. From left to right, the robot use one point contact to rotate the block.}
    \label{fig:experiment_drawing_1}
\end{figure}

Next we illustrate how our method works with a concrete example. Consider the ``block tilting'' task shown in Fig. \ref{fig:experiment_drawing_1}. The robot hand is a point. The robot needs to tilt and flip a square block about one of its edges by pressing on the block's top surface. We use a simple motion plan: the robot hand moves along an arc about the rotation axis, and all contacts in the system are sticking. If we only use velocity control to execute the plan, the robot can easily get stuck since the modeling or perception of the block may not be perfect.

\subsection{Variables} % (fold)
\label{sub:variables}
\begin{figure}[h]
    \centering
    \includegraphics[width=0.4\textwidth]{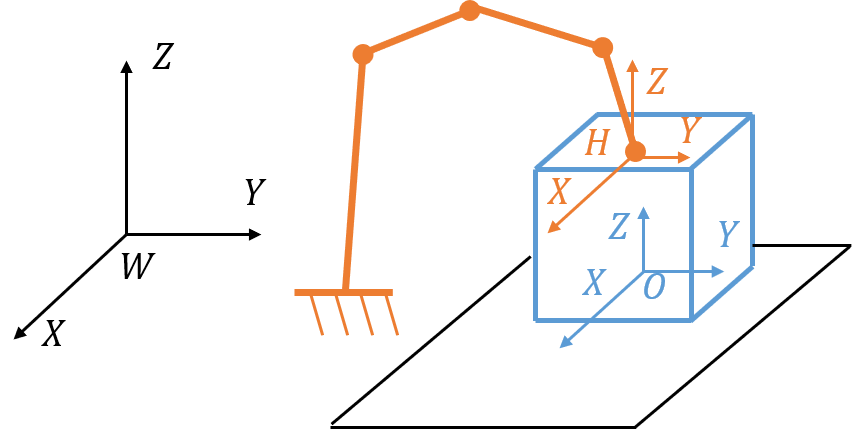}
    \caption{illustration of the coordinate frames.}
    \label{fig:experiment_drawing_2}
\end{figure}
Denote $W$, $H$ and $O$ as the world frame, the hand frame and the object frame respectively. In the following, we use the form of $^A_BX$ to represent a symbol of frame $B$ as viewed from frame $A$. We do Cartesian control for the robot, so we ignore the joints and only model the hand. The state of the system can be represented by the 3D pose of the object and the position of the hand as viewed in the world frame:
\begin{equation}
	{q}=[^W_Op^T,\ ^W_Oq^T,\ ^W_Hp^T]^T\in \mathbb{R}^{10}.
\end{equation}
Define the generalized velocity for the system to be the object body twist $^O_O\xi\in \mathbb{R}^6$ and the hand linear velocity $^W_Hv\in\mathbb{R}^3$:
\begin{equation}
	v=[^O_O\xi^T,\ ^W_Hv^T]^T\in\mathbb{R}^9.
\end{equation}
The benefit of choosing body twist over spatial twist for representing generalized velocity of rigid body is that the expression of the mapping $\dot q=\Omega(q) v$ becomes simple:
\begin{equation}
\Omega ({\bf{q}}) = \left[ {\begin{array}{*{20}{c}}
{{^W_OR}}&{}&{}\\{}&{E({^W_Oq})}&{}\\{}&{}&{{I_H}}\end{array}} \right]\in \mathbb{R}^{10\times 9},
\end{equation}
where $^W_OR\in SO(3)$ denotes the rotation matrix for $^W_Oq$, $E(^W_Oq)$ is the linear mapping from the body angular velocity to the quaternion time derivatives \cite{graf2008quaternions}:
\begin{equation}
E(^W_Oq) = \frac{1}{2}\left[ {\begin{array}{*{20}{c}}
{ - {^W_Oq_1}}&{ - {^W_Oq_2}}&{ - {^W_Oq_3}}\\
{{^W_Oq_0}}&{ - {^W_Oq_3}}&{{^W_Oq_2}}\\
{{^W_Oq_3}}&{{^W_Oq_0}}&{ - {^W_Oq_1}}\\
{ - {^W_Oq_2}}&{{^W_Oq_1}}&{{^W_Oq_0}}
\end{array}} \right].
\end{equation}
The generalized force corresponding to our choice of generalized velocity is the object body wrench together with the hand pushing force:
\begin{equation}
	f=[^O_Ow^T,\ ^W_Hf^T]^T \in\mathbb{R}^9
\end{equation}
\subsection{Goal Description} % (fold)
\label{sub:example_goal_description}
In the motion plan, the object rotates about the line of contact on the table. The goal for control at any time step is to let the object follow this motion. Now we try to write down the generalized velocity for such motion.
Denote $^Wp_{tc}$ as the location of any point on the line of contact, $^W\omega_g$ as the axis of rotation, $\dot\theta_g$ as the desired object rotation speed. We can firstly write down the spatial twist for the object motion as $^W\xi_g=(-^W\omega_g\times ^Wp_{tc},\ ^W\omega_g)\dot\theta_g\in\mathbb{R}^6$. The corresponding body twist can be computed as
\begin{equation}
^O\xi_g=Ad_{^W_Og^{-1}}{^W\xi_g}
\end{equation}
where $Ad_{^W_Og^{-1}} = \left[ {\begin{array}{*{20}{c}}{^W_OR^T}&{ - ^W_OR^T{^W_O{\hat p}}}\\0&{^W_OR^T}\end{array}} \right]$
is the adjoint transformation associated with $^W_Og^{-1}=\left[ {\begin{array}{*{20}{c}}{{^W_OR}}&{{^W_Op}}\\0&1\end{array}} \right]^{-1}$. Then the goal for our controller can be specified as
\begin{equation}
\label{eq:example_goal}
G{\bf v}=b_G,
\end{equation}
where $G = \left[ {\begin{array}{*{20}{c}}{{I_6}}&0_{6\times3}\end{array}} \right],\;\;\,{b_G} = {^O\xi_g}.$

\subsection{Natural constraints} % (fold)
\label{sub:example_natural_constraints}

\subsubsection{Holonomic constraints} % (fold)
\label{subsub:example_holonomic_constraints}
The contact between the object and the hand  is a sticking point contact,  which constrains the system states by
\begin{equation}
\label{eq:hand_contact_holonomic}
^W_OQ(^Op_{hc}) + ^W_Op = ^Wp_{hc},
\end{equation}
where $^Wp_{hc}, ^Op_{hc}$ denote the location of the contact point, function $^W_OQ(p)$ rotates vector $p$ by quaternion $^W_Oq$.

The contact between the object and the table is a sticking line contact. We approximate it with two point contacts at the two ends. Use subscript $tc$ to denote the table contacts, the sticking constraints can be approximated by requiring the two points to be sticking:
\begin{equation}
\label{eq:table_contact_holonomic}
{{^W_OQ}{(^O}{p_{tc,i}}) + {^W_Op}} { = }{^Wp_{tc,i}},\;\;\; i=1,2.
\end{equation}
Equation (\ref{eq:hand_contact_holonomic}) and (\ref{eq:table_contact_holonomic}) together form the holonomic constraints for our system:
\begin{equation}
\label{eq:example_holonomic_constraint}
\Phi ({{q}}) = \left[ {\begin{array}{*{20}{c}}
{^W_OQ(^Op_{hc}) + ^W_Op = ^Wp_{hc}}\\
{{^W_OQ}{(^O}{p_{tc,1}}) + {^W_Op}} { = }{^Wp_{tc,1}}\\
{{^W_OQ}{(^O}{p_{tc,2}}) + {^W_Op}} { = }{^Wp_{tc,2}}\end{array}} \right] = 0
\end{equation}
This example does not have face to face contacts; they can be handled similarly by multi-point-contacts approximation.

\subsubsection{Newton's second law} % (fold)
\label{subsub:example_newton_s_second_law}
The reaction forces $\lambda=[^W\lambda_{hc}^T,^W\lambda_{tc,1}^T,^W\lambda_{tc,2}^T]^T\in \mathbb{R}^9$ associated with the holonomic constraints (\ref{eq:example_holonomic_constraint}) are the three contact forces as viewed in world frame. In Newton's second law (\ref{eq:newton_second_law}):
	 $$\Omega^T(q)J^T_\Phi(q)\lambda + f + F = 0$$
$\Omega$ is known, $J_\Phi(q)$ is computed by symbolic derivation from $\Phi(q)$, we refrain from showing its exact expression to save pages. The external force $F$ contains the gravity of the object $G_O$ and the robot hand $G_H$, the reference frames of which should be consistent with the generalized force:
\begin{equation}
F = \left[ {\begin{array}{*{20}{c}}{^O{G_O}}\\
0\\{^H{G_H}}\end{array}} \right] \in\mathbb{R}^9.
\end{equation}
$^HG_H$ should be zero if the robot force controller already compensates for self weight.
% subsection newton_s_second_law (end)

\subsection{Guard Conditions}
\label{sub:example_guard_conditions}
The motion plan requires all contacts to be sticking. Coulomb friction thus gives two constraints on the force variables:
\begin{enumerate}
	\item The normal forces at all contacts must be greater than a threshold $n_{min}$.
	\item All contact forces must be within their friction cones.
\end{enumerate}
To express 3D friction cone constraints linearly, we approximate the cone with eight-sided polyhedron \cite{trjopt1} with $d_i=[\sin(\pi i/4),\ \cos(\pi i/4),\ 0]^T$ being the unit direction vectors for each ridge. Denote $\mu_{hc}, \mu_{tc}$ as the estimated minimal possible friction coefficient, $z=[0\ 0\ 1]^T$ as the unit $Z$ vector, the friction cone constraints becomes
\begin{equation}
\label{eq:example_friction_cone}
	\begin{array}{l}
{\mu _{hc}}{z^T}(_W^OR{^W\lambda _{hc}}) \ge d_i^T(_W^OR{^W\lambda_{hc}}),\;\;\,\;\;\,i = 1,...,8\\
{\mu _{tc}}{z^T}{^W\lambda _{tc,1}} \ge d_i^T{^W\lambda _{tc,1}},\;\;\,\;\;\,i = 1,...,8\\
{\mu _{tc}}{z^T}{^W\lambda _{tc,2}} \ge d_i^T{^W\lambda _{tc,2}},\;\;\,\;\;\,i = 1,...,8
\end{array}
\end{equation}
The normal force lower bound can be written as
\begin{equation}
\label{eq:example_normal_force}
	\begin{array}{l}
	{z^T}({_W^OR}{^W\lambda _{hc}}) \ge {n_{\min }}\\
	{z^T}^W{\lambda _{tc,1}} \ge {n_{\min }}\\
	{z^T}^W{\lambda _{tc,2}} \ge {n_{\min }}
	\end{array}
\end{equation}
Equation (\ref{eq:example_friction_cone}) and (\ref{eq:example_normal_force}) are affine constraints on $\lambda$, together they form the guard condition (\ref{eq:guard_conditions}).

\subsection{Solve the problem} % (fold)
\label{sub:example_solve}
At each time step, given the object and the hand poses we can use algorithm \ref{alg:solve_for_velocity} and \ref{alg:solve_for_force} to solve for the hybrid force-velocity control numerically. You can find our Matlab implementation of the step by step derivations in our GitHub repository (see section \ref{sub:code}).

Here we briefly describes the solved actions. The solution to the block tilting problem has one dimensional velocity controlled action, which points in the tilting direction and is roughly perpendicular to the line from the hand to the rotation axis. The other two dimensions are under force control. The $Y$ component of the force command is close to zero, which makes sense as forces in $Y$ direction don't do anything useful. The force in other component is roughly pressing against the rotation axis to maintain sticking.

% !TEX root = ../ICRA19Hybrid.tex

\section{EXPERIMENTS} % (fold)
\label{sec:experiments}
\begin{figure}[h]
    \centering
    \includegraphics[width=0.5\textwidth]{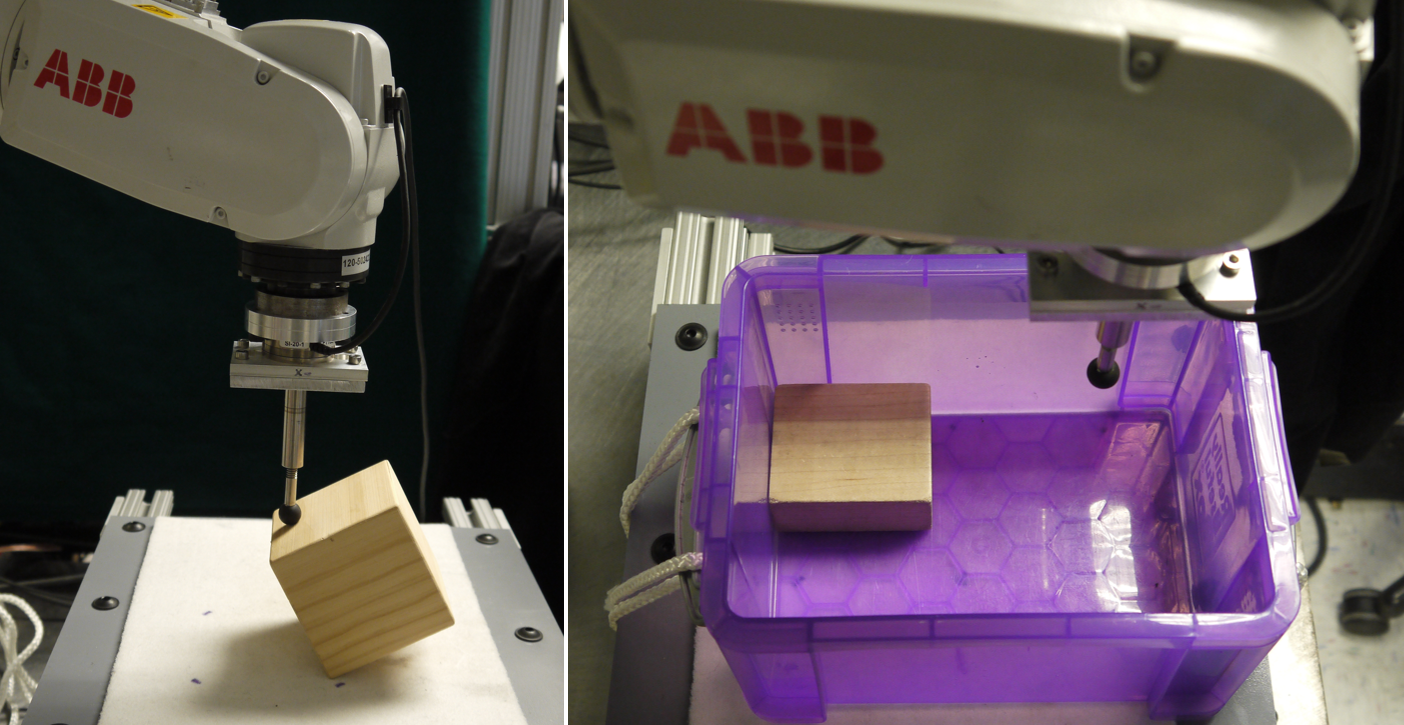}
    \caption{Our experiment setup. Left: block tilting. Right: tile levering-up.}
    \label{fig:experiment_photo_1}
\end{figure}
We implemented two example tasks: block tilting (section~\ref{sec:example}) and tile levering-up. In the tile levering-up example, the robot need to pivot the object up against a corner in a box. The object is modeled as a cuboid. During the motion, the contacts between the object and the corner are sliding, while the contact between the object and the robot is sticking.

In the block tilting task, the object is a wooden block with edge length 75mm. We place a 2mm-thick piece of cloth on the table to introduce some passive compliance as well as increasing friction. In the tile levering-up example, the object is placed at a corner of a plastic box, which is fixed in space. We experimented with a variety of objects.
The robot hand is a metal bar with a rubber ball installed on the tip to increase friction.

We implemented our algorithm \ref{alg:solve_for_velocity} and \ref{alg:solve_for_force} in both Matlab and C++. The projected gradient descent is the most time-consuming part of our algorithm. With $N_s = 3$ initial guesses, the C++ code can solve the block tilting problem in 35ms, solve the levering-up problem in 25ms. The Matlab version is on average 10x slower.

The control computed by our algorithm can be implemented in many ways. We implemented hybrid force-velocity control with position-control inner loop according to \cite{maples1986experiments}, and added functionality for choosing axes in any orientation. We used an ABB IRB 120 robot arm with 250Hz communication (but with 25ms latency), and a wrist-mounted force torque sensor, ATI Mini-40, to measure contact forces at 1000Hz.

We ran the block tilting task 50 times in a row\footnote{You can find the 25min video at \href{https://youtu.be/YIP8xIFATHE}{https://youtu.be/YIP8xIFATHE}}. Each run contains 15 time steps. The robot successfully tilted the block 47 times. The three failures were all stopped prematurely because the robot detected large force (about 25N on the FT sensor) at a time step. The reason could be a bad solution from our algorithm, or the instability of our force control implementation.

We ran the tile levering-up task for about 20 times on different objects. The successful rate is about two-thirds. The failures are caused by unexpected sticking between the object and the wall, or unexpected slipping between the robot hand and the object. One important reason for these failures is the slow response of the low level force control: the commanded positive contact normal force were tracked with large errors, which could surely be improved with better engineering. We did observe that the failures are less likely to happen if the robot moved slower.

\subsection{Resources} % (fold)
\label{sub:code}
The Matlab implementation of the two algorithms along with several examples can be obtained from \href{https://github.com/yifan-hou/pub-icra19-hybrid-control}{https://github.com/yifan-hou/pub-icra19-hybrid-control}.
 % You can also download our implementation of the low level hybrid force-velocity controller from \href{https://github.com/yifan-hou/forcecontrol}{another GitHub repository}.

% !TEX root = ../ICRA19Hybrid.tex

\section{DISCUSSION AND FUTURE WORK} % (fold)
\label{sec:discussion_and_future_work}
For a hybrid force-velocity control problem, people might be able to manually design a control strategy that works just fine. We insist that our method is valuable, because we can automate the process for new problems without manual design. Moreover, we can solve some problems that are unintuitive for a human.
\begin{figure}[ht]
    \centering
    \includegraphics[width=0.35\textwidth]{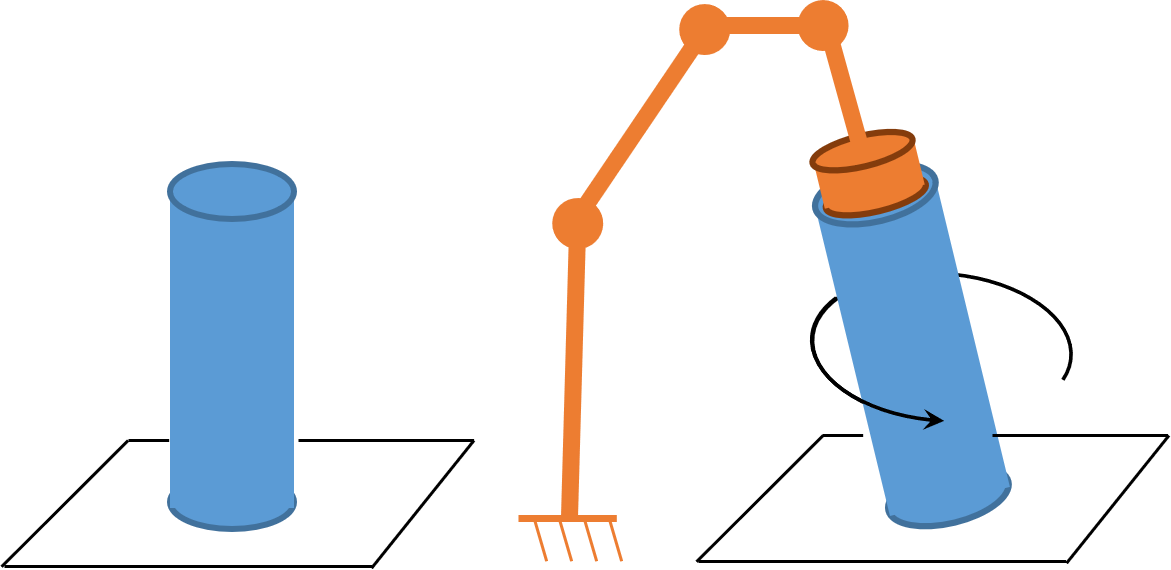}
    \caption{Illustration of the bottle rotation problem.}
    \label{fig:experiment_drawing_3}
\end{figure}
For example, consider the bottle rotation problem as shown in Fig. \ref{fig:experiment_drawing_3}, left. Use a robot to press on its top surface with a face to face contact. If you apply force properly, you can tilt the bottle and rotate it on the table. The control strategy is not straightforward, since it involves hybrid actions in 6D wrench space. The Matlab code for solving this problem is also available in our GitHub repository. Unfortunately we don't have time to implement it on a robot.

Our method has several limitations. Firstly, we haven't consider non-holonomic constraints in our current formulation. Secondly, the algorithm \ref{alg:solve_for_velocity} could get stuck in a bad local minimum. The only way of avoiding it is to sample more initial points, which increases computation time. Finally, the cost functions proposed in this work are largely based on our intuition. The exact conditions for maintaining contact modes are not completely clear. Although they seem to work empirically, a better understanding of contact mechanics may lead to a more reliable hybrid servoing algorithm.

% !TEX root = ../ICRA19Hybrid.tex

\section*{ACKNOWLEDGMENT}
We would like to thank Hongkai Dai, Nikil Chavan-Dafle and Fran{\c{c}}ois Hogan for helpful discussions and suggestions.

\bibliographystyle{plain}
\bibliography{ICRA19Hybrid}

\end{document}